\documentclass{article}

\usepackage[utf8]{inputenc}
\usepackage{arxiv}

\usepackage{amsmath}
\usepackage{hyperref}
\usepackage{url}
\usepackage{graphicx}
\usepackage{xcolor}
\usepackage{booktabs}
\usepackage{lipsum}
\usepackage{enumitem}
\usepackage{float}
\usepackage{array}
\usepackage{tablefootnote}
\usepackage{comment}
\usepackage{hyperref}
\usepackage{amsfonts}       % blackboard math symbols
\usepackage{nicefrac}       % compact symbols for 1/2, etc.
\usepackage{microtype}      % microtypography

\usepackage[T1]{fontenc}
\usepackage[utf8]{inputenc}

\usepackage[numbers]{natbib}
\usepackage{doi}

\usepackage{makecell}
\usepackage{comment}

\usepackage{xurl}

\definecolor{fig3pink}{RGB}{255, 66, 161}
\definecolor{fig3green}{RGB}{0, 108, 101}
\definecolor{fig3blue}{RGB}{117, 177, 212} % This color works well.
\definecolor{fig3purple}{RGB}{128, 12, 70}

\title{Towards cross-cultural study of folksong lyrics with machine translation}
\author{%
Anna Dvořáková \\ Charles University \\ Czech Republic \And %
~Anna Aljanaki  \\ University of Music and Performing Arts Graz \\ Austria \And %
~Danbinaerin Han \\ Graduate School of Culture Technology \\ KAIST \\ Republic of Korea \And %
~Peter van Kranenburg \\ Utrecht University \\ Netherlands \And %
~Matěj Kratochvíl \\ Institute of Ethnology \\ Czech Academy of Sciences \\  Czech Republic \And %
~Inna Lisniak \\ Estonian Literary Museum; \\ M. T. Rylsky Institute of Art Studies, \\ Folkloristics and Ethnology \\ NAS of Ukraine \\  \And %
~Zdeněk Vejvoda \\ Institute of Ethnology \\ Czech Academy of Sciences \\  Czech Republic \And %
~Jan Hajič jr.* \\ Charles University \\ Czech Republic \\ \texttt{hajicj@ufal.mff.cuni.cz}
}

\date{}

\begin{document}

%%%%%%%%%%%%%%%%%%%%%%%%%%%%%%%%%%%%%%%%%%%%%%%%%%%%%%%%%%%%%%%%%%%%%%%%%%%%%%%%
% Abstract
%%%%%%%%%%%%%%%%%%%%%%%%%%%%%%%%%%%%%%%%%%%%%%%%%%%%%%%%%%%%%%%%%%%%%%%%%%%%%%%%

\twocolumn[{%
\maketitle
\begin{abstract}
Music is universally present in human societies. 
Ethnomusicologists have long been 
documenting the diverse expressions of human musicality, and comparative musicology has recently brought several studies of folksong to a more global scale. 
Such cross-cultural research has not been conducted on lyrics:
the language barrier has so far prevented work with multi-lingual data.
However, Natural Language Processing (NLP) technologies have reached a stage where this language barrier may no longer be prohibitive.
Combining folksong lyrics corpora across five languages, %we machine-translate them to a pivot language, represent them using a pre-trained neural topic model, 
we machine-translate them to a pivot language with a pre-trained neural topic model, and we examine the relationship between content and social function within each language, and across languages for wedding songs.
As expected, human evaluation of translation results shows that 
non-Indo-European languages suffer from overall worse translation quality. 
Experiments with topic models then indicate that the content of lyrics is at best partially related to the social function of folksongs across all languages.
These experiments are just first steps into cross-cultural folk musics lyrics analysis; however, they do indicate that a previously unobserved web of cross-cultural relationships beyond ethnomusicological typologies may be uncovered through the study of what people sing across the world's diverse folk musics.
%
%examine the relationship between their content and social function --- wedding songs, lullabies, work songs, romances, and more --- using a pre-trained neural topic model. The results show that the relationship between content and social function 
%While these results should not be taken as more than 
%and as a case study we examine what is shared and what is not in the textual content of wedding songs is shared and different across the five musical cultures.
%however, no dataset of folk song lyrics from diverse cultural backgrounds exists. We address this resource gap (1) by collecting a global dataset of folk music lyrics from existing corpora for individual cultures, (2) by making them comparable across the language barrier using multilingual representations and machine translation capabilities of Large Language Models, and (3) by proposing a shared typology, cross-checking theories of anthropological universals and ontologies of individual corpora with structures discovered in the raw data using topic models. We are confident this dataset will be a valuable resource for understanding what people sing around the globe.
\end{abstract}
\hspace{24pt}%
}]
%\saythanks{}

%%%%%%%%%%%%%%%%%%%%%%%%%%%%%%%%%%%%%%%%%%%%%%%%%%%%%%%%%%%%%%%%%%%%%%%%%%%%%%%%
% Main Content Start
%%%%%%%%%%%%%%%%%%%%%%%%%%%%%%%%%%%%%%%%%%%%%%%%%%%%%%%%%%%%%%%%%%%%%%%%%%%%%%%%

%%%%%%%%%%%%%%%%%%%%%%%%%%%%%%%%%%%%%%%%%%%%%%%%%%%%%%%%%%%%%%%%%%%%%%%%%%%%%%%%%%%%%%%%%%%
\section{Introduction}
\label{sec:introduction}

Every society sings \cite{mehr_universality_2019}. The diversity of musics across societies is immense \cite{list_nonuniversality_1971}, and the way societies are related through shared musical practices is globally independent of linguistic or genetic similarities between populations \cite{passmore_global_2024}, making music a distinct layer at which relationships between cultures are embodied, and justifying the study of music across cultures, termed a ``new comparative musicology'' \cite{savage_comparative_2025}. 
Because of how the field studying musical diversity has historically constituted itself (long before globalisation), the materials of this study have mostly been termed folk musics, with the Cantometrics project as the most prominent early example of global ambition \cite{lomax_cantometrics_1978,savage_alan_2018}.

Practically all this cross-cultural research, however, has focused on \textit{how} people sing --- the musical features \cite{lomax_cantometrics_1978,wood_global_2022,ozaki_globally_2024} and building blocks \cite{McBride_convergent_2023,jacoby_commonality_2024}. 
Some has focused on \textit{why} people sing \cite{mehr_universality_2019,savage_many_voices_2025}, and recently on how people feel based on music \cite{lee_globalmood_2025}.
%The features of Cantometrics and the Global Jukebox range from type of polyphony, phrase lengths, vocal techniques used, all the way to the social organisation of musicking groups \cite{lomax_cantometrics_1978,wood_global_2022}. Cross-cultural studies of both musical scales \cite{McBride_convergent_2023} and rhythmic ratios \cite{jacoby_commonality_2024} show that these musical ``building blocks'' are remarkably consistent globally. 
%Global musical diversity has also been studied in connection with \textit{why} people sing: behavioural categories \cite{mehr_universality_2019}, showing that music varies among three main dimensions (formality, arousal, religiosity) within societies more than across them. The Many Voices project identified globally relevant distinctions between song and speech audio across 75~languages \cite{ozaki_globally_2024}, and may show that singing is uniquely powerful at building pro-social attitudes \cite{savage_many_voices_2025}, and GlobalMood surveyed perceived emotion in music across 59 countries \cite{lee_globalmood_2025}.
%
But none of this research foregrounds what people sing about: the lyrics. 
But the text of folksongs might matter a great deal. If music, especially folk music, is understood as a mechanism of constructing and affirming identity \cite{firth_music_identity_1996,rice_reflections_identity_2007} and group cohesion \cite{savage_many_voices_2025}, then the way language is used in music could provide a unique honest window into what concepts mean across different cultures, and therefore how cultures share worldviews (or don't). 

%and for over two centuries collecting folk songs (lyrics and melodies) has been a key method of ethnography, thus there exist national corpora of folk music and some of them even have digital form (see section \ref{sec:methods}). 
%"Why is there no comparative work on lyrics?"
%
The avoidance of lyrics is no mystery, especially when computational methods are involved. Pitches \cite{McBride_convergent_2023}, rhythms \cite{jacoby_commonality_2024}, and derived musical features can be somewhat directly compared across cultures, but for cross-cultural work on lyrics, language differences pose a major barrier. %(as music is one language)
However, advances in machine translation might be making it feasible to work across languages.
%that have so far precluded cross-cultural work on folk music lyrics (as opposed to melodies or other musical properties). 
We believe this research direction is therefore worth exploring. Providing an impulse for this work with a specific experiment, and mapping the issues this research direction will likely present, is the overarching objective of this paper.

%The methodology for this work are not at all clear.
%It is a research direction fraught with methodological danger. 
%There is much anthropological and sociological discussions that the cross-cultural study of sung texts will require.  Most of all, 
It is hard (some might argue impossible) to compare what song texts ``mean'' to members of two different societies and cultures. But there are things we can do without having to resolve such debates.
Specifically, \textbf{we can observe what cultures prioritise to express in individual \textit{social contexts}} in which their music is performed:
weddings, lullabies, work songs, funeral songs, carols, or other ``types'' of folksong. 
Classifying folksong by the social contexts to which they belong --- at least, in which they were observed --- has been a staple of ethnomusicological collection activity.
%
%There are even ``anthropological universals'' \cite{brown_human_2004} that could serve as a basis for mapping the social contexts of one society to the other, although for now, 
%cross-cultural research into the relationship of folksong lyrics and their social context is 
%we are limited to the typologies as they were defined by individual ethnomusicological collection efforts. (Before effort is invested into broader discussions about harmonising these typologies, the feasibility of crossing the language barrier should be investigated first.)
%
So, a first step in the direction of cross-cultural research on folksong lyrics can be to 
%gather resources %and tools and 
perform exploratory experiments relating them to social contexts.
%so that at the very least there is a concrete object that can be subjected to criticism, stimulating these necessary methodological discussions.

We gather folksong lyrics corpora in Dutch, Czech, Ukrainian, Estonian, and Korean, which all have metadata denoting social contexts of individual songs. We project the lyrics into a shared space by translating them into English as a pivot language, 
and using the usual topic model approach 
%(which has been most prevalent in folksong lyrics research so far 
\cite{strle_language_2016,petrovych_topic_2024,grootendorst2022bertopic}, we ask:
\textbf{Is there even a relationship between what is sung and the social context in which a song is sung?}
Just looking at the lyrics, is it possible to tell whether a song is an agricultural work song? A lullaby? And then: by looking at the text of, e.g., a wedding song, is it possible to guess which culture it comes from? %Which cultures' wedding songs might get most confused? %and what is it that multiple cultures sing about when a wedding is involved? 
%The background plan is to establish

We find that machine translation performs well despite the presence of dialect, though it degrades in presence of non-standard orthography. But across all languages, the relationship between lyrics content and social function is complicated, and concepts that each culture chooses to express in individual social contexts are distinct enough to classify only 2 out of 3 song texts correctly. Across wedding songs, there are considerable differences --- but at the same time the Korean control group is not at all classified perfectly. Considered together, we believe this research space is worth further exploration. 

In the rest of this paper, 
we briefly review related work (Sec.~\ref{sec:relatedwork}), 
introduce the five corpora (Sec.~\ref{sec:corpora}),
describe the choice of methods (Sec.~\ref{sec:method}, both for machine translation and then topic modeling),
and finally report results (Secs.~\ref{sec:mt} and \ref{sec:results_class}) and discuss the findings (Sec.~\ref{sec:conclusions}). All code and data are available in Supplementary materials.\footnote{\url{https://anonymous.4open.science/r/ISMIR26_Folksong_weddings-B7F5}}

%Much coarser-grained goal: figuring out the relationship between content and function, or social context. What is sung vs. why is it sung. This has been done for musical properties of folksongs: patterns emerge where some function has cross-culturally shared musical properties (lullabies), but other types of song cannot be "guessed" this way.

%\textbf{Note:} we need to define ``function'' or rather ``social context'' here. Relationship to Mehr et al., anthropological universals, but in practice we are restricted for now to typologies developed by ethnographers for individual folksong corpora. (Also, the need for typologies restrict us to corpora that even have a typology.)
%
%Comparing function (social context?) is an anthropological problem. Some ethnomusicological traditions have produced typologies of folksong. These typologies do not necessarily map onto each other perfectly, but they also are not completely incomparable. There are certain ``anthropological universals'' that can serve as bridges, or rather anthropologically validated dimensions of shared spaces.

\section{Related work}
\label{sec:relatedwork}

Cross-cultural research on folk music is exemplified by Alan Lomax's Cantometrics project \cite{lomax_cantometrics_1978, savage_alan_2018}, which proposed ten major ``song-style regions'' based on features such as social organisation of performing groups, phrase lengths, polyphony, and vocal technique, recently refined in the Global Jukebox \cite{wood_global_2022}. 
Music was shown to vary in formality, arousal, and religiosity more within societies than across them \cite{mehr_universality_2019}. 
The Voices of Civilizations project \cite{wu_voices_2025} proposes to work with audio recordings across 38 languages; 
the Many Voices project identified globally relevant acoustic distinctions between song and speech across 75 languages \cite{ozaki_globally_2024}.
GlobalMood surveyed perceived emotion in music across 59 countries \cite{lee_globalmood_2025}.
More detailed work also focused on melodic patterns
\cite{brinkman_cross-cultural_2021, juhasz_low_2011, juhasz_systematic_2006, savage_alignment_2022}.

% Mela by začít "hlavně melodie" část
These cross-cultural studies build on extensive prior ethnographic work. 
Collecting lyrics, melodies, and other musical properties of folk songs and their performance has been a central part of ethnography for over two centuries, and continues in the digital domain \cite{miklos_polyphony_2025, sierra_recording_2023, shanahan_densmore_2014}, resulting in numerous ``national'' corpora at varied stages of digitisation (the corpora used in the present study are described in Sec.~\ref{sec:corpora}). 
% \textbf{All the cross-cultural studies focus overwhelmingly on the musical features of folk musics and not on the lyrics.} e.g. \cite{brinkman_cross-cultural_2021, juhasz_low_2011, juhasz_systematic_2006} 
%
Computational work within single cultures also focused on musical features: melodic analysis of Hungarian \cite{juhasz_variation_2000}, %\cite{juhasz_structure_2002, juhasz_systematic_2006}, 
Portuguese \cite{carvalho_computational_2023, bernardes_modal_2024}, melody pattern matching on Dutch \cite{volk_melodic_2012}, Basque \cite{conklin_antipattern_2013}, Slovenian \cite{borsan_adding_2023}, and Native American \cite{neubarth_supervised_2018} folk songs, as well as emotion classification from audio features of Chinese folk songs \cite{li_research_2022}.

NLP methods have been applied to song lyrics more broadly \cite{hunke_evolution_2025, brand_cultural_2019}, but work on folk song lyrics remains largely confined to single cultures: Korean \cite{han_finding_2023}, Chinese \cite{li_research_2022}, with classification of folk songs using emotional words in lyrics, or Ukrainian and Slovenian in the topic analysis of folksongs \cite{petrovych_topic_2024, strle_language_2016}.
%
%Much more focus of ``new comparative musicology'' has been directed toward ``how'' people sing than ``what'' they sing.
%
%Lyrics: that Hungarian guy with a multi-language dataset that isn't available?
%
A promising exception is an ongoing %digital folkloristics 
project %at the Estonian Literary Museum, which applies 
applying similar methods to compare 
%LLM-based translation, topic modelling, and 
alignment with folkloristic traditions 
on Ukrainian and Estonian corpora \cite{petrovych_oral_2025, petrovych_modeling_mythological_2026}.
%, though it has yet to produce results.

%petrovych_topic_2024}.
%
% These technologies would enable taking a similarly broad look on lyrics, next to melodies and other musical features: a look that would include multiple cultures and ask more general questions.
% %like how does folk music material align with existing universals 
% These include alignment of folk music material with established cultural universals (general \cite{brown_human_2004} or musical \cite{savage_comparative_2025}) or other proposed typologies from the domain of ethnomusicology. 
% %In such a case, collaboration with other fields (such as ethnology or anthropology) is necessary.
% % (1) What underlying themes, motifs, and narrative structures can be identified within Ukrainian and Estonian folk songs using topic modelling? 
% % (2) How do the thematic structures derived from computational analysis align with traditional folkloristic classifications? 
% % (3) How does the use of translation impact the analysis of thematic overlap between these two languages?
% %

% ~~~~~~~~~~~~~~~~~~~~~~~~~~~~~~~~~~~~~~~~~~~~~~~~~~~~~~~~~~~~~
\section{Corpora}
\label{sec:corpora}

We use folksong lyrics from five languages: Dutch (NL), Czech (CS), Ukrainian (UK), Estonian (ET), and Korean (KO). The selection was made based on the availability of a typology of social contexts within the respective corpus.
%which is what enabled the identification of wedding songs.
In further text, we call the individual typologies the ``labels''.
%
%In the following sections, we present individual corpora. 
Tab.~\ref{tab:corpora} specifies sizes 
%an overview of the sizes 
of individual corpora.

% Please write here your paragraph about "your" corpus:
% - Origin
% - Size (no. of songs, no. of lyrics words) - Anna D. will add
% - One or two sentences about the origin of the song typology used, if possible. 
% - The citation (goes into the *.bib file)

\subsection{Dutch Song Database (NL)}

% Peter
% Dutch Song Database

The Dutch Song Database\footnote{\url{https://www.liederenbank.nl}} \cite{van-Kranenburg2019} is an online accessible database that contains references to Dutch songs in historical sources (mainly printed songbooks, manuscripts, songsheets, and field recordings). It is hosted by the Meertens Institute in Amsterdam. The database started in the early 1990s from the work on 17th-century Dutch song culture by the late Dutch musicologist Louis Peter Grijp. 
Most of the 17th century sources do not include music notation, but refer to the intended melody by a tune indication that was readily understandable to contemporary singers. These tune indications were not standardized. The same tune could be referred to by different names, or variants of the same name. The initial aims of the Dutch Song Database were to have a searchable system that would connect all tune indications belonging to a given tune, and to retrieve all known tunes that fit a given set of lyrics based on their metric profile.

Through a series of projects, the database has been expanded to its current size of c.\ 180 thousand song references. 
For this study, we restricted ourselves to songs from the 18th, 19th and 20th centuries.
Virtually all known sources up until the 16th century have been included. From the 17th century on, the coverage is more selective. Both instrumental and vocal sources have been incorporated as they share the same repertoire of melodies. For the 20th century, the main corpus of songs consists of field recordings that were made by Dutch ethnologists Will Scheepers and Ate Doornbosch during the 1950s till 1980s. Next to metadata, the database includes full lyrics for c.\ 50 thousand songs, and digitally encoded melodies for c.\ 20 thousand songs.

Over its life span of more than three decades, the database has been continuously enriched and updated by more than one hundred collaborators, including collection specialists, interns, students, and volunteers. 
A particular relevant enrichment for the current study is the assignment of keywords to the songs. Over the years, this effort has developed into a reasonably coherent taxonomy consisting of c.\ 250 keywords in 15 categories.

% Anna D:
% Subset is used - 18th, 19th and 20th century, only some genres: 
% bruiloftslied, begrafenislied, verjaardagslied, ambachtslied, zeemanslied, soldatenlied, wiegelied, schommellied, driekoningenlied, sinterklaaslied, kinderlied, anti-huwelijkslied
% This search list will be present in the supplementary repository

\subsection{Písňovna (CS)}

% Zdeněk V. and Matěj K.
% Project by Ethnological Institute of Czech Academy of Science

The corpus of Czech folk songs is derived from data from the Písňovna portal,\footnote{\url{https://pisnovna.cz/}, \url{https://badatelska.pisnovna.cz/}} which was created by the Department of Ethnomusicology and Ethnochoreology at 
the Institute of Ethnology of the Czech Academy of Sciences through the digitisation of the historical manuscript archive of musical folklore and some older printed collections 
\cite{pisnovna2025}. 
%The main authors of the portal and related databases are Matěj Kratochvíl, Petr Nuska, Zdeněk Vejvoda, Lucie Uhlíková, and Marek Janata.  
As of April 2026, Písňovna contains more than 15 thousand entries of folk songs and instrumental melodies from various regions of the Czech Republic recorded between 1819 and 1970. 
For the current analysis, we used transcriptions of Czech folk song lyrics from the period spanning the first half of the 19th century to the mid-20th century. 

This data comes from several distinct collection efforts.
A large portion of the examined material 
was collected by Karel Jaromír Erben (1811–1870) 
in 1830--1865 between 1830s and the early 1860s 
and published by him in print between 1862 and 1864. 
The collection Prostonárodní české písně a říkadla 
contains more than 2,000 
song lyrics and 811 melodies 
from all regions of the then Kingdom of Bohemia, and is organised thematically (children’s songs, ceremonial songs, love songs, dance songs, humorous songs, wedding songs, songs about professions, military songs, 
etc.) \cite{vejvoda_erben_2017}.

Another portion of the material for analysis was collected in the vicinity of Prague and in Central Bohemia by the teacher 
František Homolka (1885–1933) 
before 1918 \cite{hom_podrip_2015, hom_praha_2013, hom_praha_2011}.
His collection is a part of the archive of the Institute of Ethnology. It was created in connection with the centrally organized campaign Das Volkslied in Österreich and the subsequent activities of new institutions of the First Czechoslovak Republic. 
The repertoire of his informants reflects the pivotal period of the decline of traditional folk culture associated with the agrarian environment and the onset of industrialization in the wider Prague metropolitan area. It 
reliably reflects the full range of musical expressions of the time.
Last is the anthology of Czech folk songs from the Bohemian-Moravian Highlands %, %edited by Věra Thořová %(*1936) 
\cite{thorova_HB_2009}, whose textual material was also analytically utilised. %includes a critical edition of Božena Koutníková’s (1876–1942) from 1910 and Thořová’s own field recordings from the same region from 1958–1959. 
Both collections are thematically and functionally structured in the printed edition (epic, love, drinking, military, carols, etc.).
%
%\subsubsection{Merging collections}
%

The Czech data combines multiple collections with somewhat distinct typologies that have not yet gone through a harmonisation process on the level of database editorial policy.
During the process of merging data of those collections, 
we merged some labels across the collections as they were clearly the same label\footnote{E.g., \textit{Entertainment \& dance} and \textit{Dance \& entertainment} songs} %(e.g. \textit{ŽERTOVNÉ A TANEČNÍ PÍSNĚ} and \textit{TANEČNÍ A ŽERTOVNÉ PÍSNĚ}) 
or would possibly have a big overlap.\footnote{E.g. \textit{Military songs} and \textit{Military songs and songs about military}} %(\textit{PÍSNĚ VOJENSKÉ} and \textit{PÍSNĚ VOJENSKÉ A O VOJNĚ}). 
%Later on, we used translations of those for easier interpretation of the results.

% Maybe move this to discussion? Here, it feels a bit weird
%Other databases are also possibly merging songs from different collections and so possibly different typologies (Dutch, Ukrainian). %: for the Dutch song database, we know this.
%But the Dutch database's labels have been coordinated, while the labels for the Czech collection, which has been digitised only very recently, are simply the merging of the different categories that look similar enough based on the label names but without manual examination of individual songs.
%but without a long process of coordination with the collectors. \textbf{(Possible TODO:Run experiment on same set of songs that was selected after all merging, top 10 labels, and subsampling, but with original pre-merging labels.)}

\subsection{Podillia Corpus (UK)}
% - Volyn part included
% Inna
Folk music archives and research centres in Ukraine working with traditional music preserve large collections of manuscript sources and field recordings gathered from the late 18th to the present day \cite{dovhaliuk2016}. 
However, a substantial part of these materials remains insufficiently digitised, fragmented, and only partially accessible for systematic analysis \cite{mazurenko2025}.\footnote{This situation is partly shaped by the historical conditions under which these archives were formed, including the practices of the Soviet period (1921–1991), which involved centralised collecting, ideological selection of materials, and restricted access to archival holdings. It is further affected by more recent challenges related to the Russian war against Ukraine since 2014, and its escalation in 2022.} This situation has created a need for a digital musical corpus as a first step towards structuring these materials and enabling their use in computational research, largely driven by individual research efforts.

The Symbolic Corpus of Ukrainian Folk Songs \cite{lisniak_corpus_2025}
%was developed through a collaboration between ethnomusicologists Inna Lisniak, Liudmyla Yefremova, and data science specialist Anna Aljanaki in 2024. The corpus 
is primarily based on published collections of Ukrainian folk songs compiled and processed by Liudmyla Yefremova from archival data of the M. T. Rylsky Institute of Art Studies, Folkloristics and Ethnology of the National Academy of Sciences of Ukraine 
\cite{dmytrenko2014, yefremova2019}. For the present corpus, songs from these collections were automatically filtered according to ethnographic regions of Ukraine, while the associated metadata were expanded, unified, and systematically structured to facilitate computational analysis. The corpus 
currently 
comprises nearly 2,000 Ukrainian folk 
songs 
encoded in MusicXML %format, 
with accompanying lyrics and structured metadata. The majority of the material originates from Podillia, an ethnographic region located in the central-western part of Ukraine \cite{lisniak2024dataset}.
In particular, the corpus includes songs from 
Central Podillia (Vinnytsia oblast) and Western Podillia (Khmelnytsky oblast). 
The future development of the corpus aims to expand its regional coverage by incorporating song materials from other regions of Ukraine.

The corpus is organised into three main categories: ritual songs, non-ritual songs, and children’s songs. The ritual group includes 
%a cycle of songs, 
calendar-ritual (winter, spring, summer), and family-ritual (wedding, funeral). The non-ritual group comprises a diverse range of genres: ballads, lyrical, historical, humorous, and dance songs, romances, etc. Children’s songs include lullabies and play songs (\textit{zabavlianky}). 
%Ukrainian wedding songs belong to a more archaic type: family-ritual songs. 

Ukrainian wedding songs belong to the archaic family-ritual tradition and occupy a central place in the wedding rite, which functions as a structured rite of passage combining singing, movement, and symbolic ritual actions \cite{klymenko2016ua}. The defining features of Ukrainian wedding songs include the use of ritual lexicon with specific semantic and cultural significance. 
%which is often inadequately rendered in automatic translation. While translation generally conveys the plot and basic semantics, it systematically loses key features of the original, including formulaic repetition, affective meaning encoded in diminutive forms  \emph{kalynonka}(little viburnum), and culturally specific terms, which are often replaced by Western European analogues \emph{pan}(Lord). In particular, ritual vocabulary is frequently reduced \emph{korovai}(wedding bread), resulting in the loss of ethnographic specificity and symbolic meaning.

\subsection{Estonian Runosongs' Database (ET)}

Estonian Runosong (regilaul) is an ancient song tradition characteristic of Finnish, Estonian and Karelian people. This study takes a subset of songs from Järvamaa region from Estonian runosong database.\footnote{\url{https://www.folklore.ee/regilaul}} These songs were usually performed in a call-and-response manner, and contain a lot of repetition. %, so it is important to take that into account when analyzing results and percentages of wrong words. 
The database has been developed in the Estonian Folklore Archives by Regilaul database workgroup. 

Despite Estonian being a small language, it is surprisingly fragmented and there are many dialects. The most notable living dialect is \emph{Võro kiil}. 
The songs in the corpus contain many dialectal words and word forms that are, in many cases, very dissimilar to standard Estonian. % from Seto, Võro, Kodavere dialects. 
As runosong dictates a strong universal rhythm, there are syllables added to words --- sometimes just for the sake of the rhythm. 
Many words are written as they sound, omitting characters and with other orthographically unorthodox choices. There are also many dialectal lexemes that are very dissimilar to standard Estonian. For instance, in Seto language some of the everyday vocabulary is borrowed from Slavic languages and is completely absent from standard Estonian. In addition, the borrowed words do not have a standardised spelling (e.g., Seto dialectal word \emph{Parapantsik} (from Russian, 
%\emph{барабанчик}) -
meaning a little drum). 

% Some Estonian words are mistranslated although they are present in normal standard Estonian. For instance, words \emph{nadu} (husband's sister) and \emph{küdi} (husband's brother) are mistranslated, probably because there is no precise English equivalent for these words.
% Anna A.

% Selection of Jarvamaa region is based on Petrovych et.al. (2025) also we take only subset of genres: 

\subsection{Finding Tori (KO)}
\label{subsec:ko}

The Finding Tori lyrics corpus \cite{han_finding_2023} 
derives from the a large-scale cultural heritage documentation project, carried out by the 
Munhwa Broadcasting Corporation (MBC) between 1989 and 1995
project documenting folk songs through field recordings across all regions of 
South Korea (1989-1995). 
The original materials were made publicly available on the MBC website in April 2022, and 
the present 
corpus was constructed by crawling the project website to extract lyrics and associated metadata, yielding a total of 15,861 songs. 
The classification system is primarily geographical.
Functional categories are assigned only where the purpose of a song is self-evident; e.g., agricultural work songs, lullabies, or songs of lament. 
In all remaining cases, the opening line of the lyrics serves as the song title.

The lyrics were transcribed and edited by scholars of Korean literature on the basis of primary documentation produced during the original field recordings. 
The guiding transcription principle \cite{mbc_urisori} was phonetic fidelity, tempered by conformity to standard grammar where possible.
%; full conventions are given in the supplementary materials. 
From the perspective of automatic translation and computational preprocessing, this corpus presents substantial challenges: 
The texts are dense with archaic vocabulary, regional dialect forms, and obsolete onomatopoeia, 
many of which are unattested in modern dictionaries. 
Dialectal spellings, rendered phonetically by convention, frequently require inference of modern equivalents on the basis of probable pronunciation.

Notably, the corpus contains no dedicated category for wedding songs. 
Songs touching on courtship are largely narrative folk songs (seosa minyo) that centre on transgressive or extraordinary events — illicit affairs, tragic love — rather than on wedding ceremonies per se. 
For this reason, we selected sijipsari-yo (lit. ``songs of living at the in-laws' house''). %as our primary material, comprising 676 songs. 
%Closest to that is sijipsari-yo (lit. ``songs of living at the in-laws' house'').
%
This genre gives voice to the hardships experienced by daughters-in-law within the Confucian patrilineal household. 
Rather than celebrating love or marriage, these songs characteristically trace a trajectory from happiness to misfortune, with marriage itself as the turning point. 
Recurrent motifs include suicide, death, and renunciation of secular life, 
while the death of a spouse or natal family members serves as a further source of suffering 
\cite{Seo2019KoFolksong}. The genre thus functioned as an emotional outlet and a form of collective catharsis for women in premodern Korean society \cite{Yu2014Folk}. 
Because these songs centre on post-marital hardship rather than wedding celebration, the Korean data serves as a contrastive case against the other four corpora: we expect these songs to be clearly separable from wedding song traditions of other genres.

% if these songs do not turn out to be at least somewhat easier to classify to their corpus of origin than wedding songs from other corpora, then something is probably wrong either with the method, or with the way the wedding label is applied in other corpora.

\subsection{Combined wedding songs}

Marriage rituals, in their many forms, belong amont prominent 
anthropological universals \cite{brown_human_2004}.
%and music tends to have a significant role in different layers of the occasion.
%: both in the ritual and, often, celebratory aspects and support for dancing. 
Despite its many forms and social implications (e.g., for gender roles), the marriage ritual is delineated clearly enough in all cultures that we do not expect to run into issues of comparing the incomparable just because they are labelled with the same term.
%: we ``know it when we see it''. In turn, we have seen that 
And in most available folksong corpora that provide metadata on social contexts, wedding songs are well-represented.

In these five corpora, a total of 1197 wedding (resp. marriage) songs have been identified.
Korean data serves as a control group, as marriage songs with a distinct content profile.
We include all 16 Estonian multi-topic songs that are also wedding songs.
The number of songs in each corpus and the number of wedding songs are given in Tab.~\ref{tab:corpora}. These numbers were reduced during translation and experiment preparation, as shown in the third column.

\begin{table}[]
    \centering
    \begin{tabular}{rrcc}
    %\toprule
         \textbf{Language} & \thead{No. of \\ songs} & \thead{No. of wedding \\  songs} & \thead{No. of songs \\ monoling. exps.} \\ 
    \midrule
         Czech & 3320 & 211 & 2829 \\
    %\midrule
         Dutch & 3041 & 149 & 2612 \\
    %\midrule
         Estonian & 4515  & 100 & 3553 \\
    %\midrule
         Korean & 15861 & 480 & 7767 \\
    %\hline
         Ukrainian & 1764 & 257 & 1585 \\
    \bottomrule
    \end{tabular}
    \caption{Number of songs with lyrics and number of wedding songs (Korean: marriage songs) in each corpus and number of songs that were used in the experiments from each corpus. The high reduction on the Korean part was caused by a large number of missing typology labels needed for the experiments}
    \label{tab:corpora}
\end{table}

% \section{Results: Baseline monolingual topic models}
% \label{sec:monolingual}

% Reporting (probably not worth the space to make a separate table):

% The optimal number of topics for individual languages: nl = , cs = , uk = , et = , ko = .

% Achieved perplexities.

% Average distance between (non-outlier) inferred topic assignments within each language.
% (Sort of ``stability of topic models themselves''.)

\section{Methods}
\label{sec:method}

%To perform experiments across the language barrier, 
We first translate the corpora to a pivot language (English) and manually evaluate the results. Then, using a neural topic model, we measure how well can songs be classified by typology (monolingual experiments) and culture of origin (wedding songs experiments).

%\textbf{TODO:The objective is out of date} The objective is to test whether multilingual technologies --- machine translation (MT) for now --- enable direct cross-cultural work on lyrics, that is, work in a shared space that minimises the impact of the language barrier.

\subsection{Machine translation}

Neither translation, nor multilingual embeddings make the projection from a language-specific to a shared space lossless.\footnote{Even disregarding the fact that just the act of translating a folksong displaces the song from its cultural context and therefore function.}
%, multilingual technologies still do not work equally well across languages, and especially in folksong, there may be difficulties with culture- and therefore language-specific phenomena, such as e.g. the term ``Kadri sandid'' in Estonian denoting carolers that has no direct equivalent. 
%
%However, lyrics analysis methods
%for analysing text content, such as topic models, 
%already operate with lossy representations of the original texts (by design, to generalise). 
%this is unavoidable if one intends to generalise across broader sets of songs. 
%Hence, a lossy %the loss of information in the projection to a shared space of translation does not necessarily imply loss of information relevant to the analytical method. 
%
It must be verified that this lossy projection into a shared space of the pivot language does not imply critical loss of information relevant to the analytical method. %Nevertheless, this must be verified.
%that we are trying to extend from monolingual to cross-cultural application contexts.
Importantly for an exploratory study such as ours, MT enables human inspection of the results of this projection from individual languages to a shared space (which multilingual embeddings do not). We thus choose MT, and evaluate translation results for each corpus by native speakers of the language who are familiar with the repertoire.%\footnote{The co-authors of the study.}

We choose English as a pivot language because it is %by far 
the best-supported language in MT, has downstream pre-trained models, %available, 
and all evaluators are proficient in it. 
Based on the results of the most recent general machine translation competition \cite{kocmi_wmt_generalMT_2025}, we choose GPT~4.1 in the version mini as the translation model (Gemini~2.5-pro performed somewhat better overall, but GPT~4.1-mini is competitive and significantly cheaper), using %The translations were done through 
the OpenRouter platform.

\subsection{Evaluating machine translation}
\label{subsec:mteval}

We run a small MT evaluation campaign with a native speaker as an annotator for each language, focusing on lexical fidelity. %. The evaluation scheme focuses on lexical fidelity. 
We do not expect MT to retain nuance; however, our ``first question'' of relating social context to concepts prioritised in lyrics is so coarse-grained that this minimum adequacy may be enough.

%Translations in each language were checked by a native speaker.
%To support the meaningfulness of the usage of machine-translated texts for analysis, we prepared human evaluation campaign as a sanity check. We were interested in the translation of the lexical content of folk songs' lyrics. 

%An annotation website was prepared. Annotators were asked to go through a set of songs and their translations with instructions to decide for each \textbf{word} in the original language, whether it is:
%    \begin{itemize}[noitemsep,nosep,leftmargin=*]
%        \item semantically important and represented in the translation
%        \item semantically important but represented incorrectly or not at all in the translation
%        \item not important for the meaning
%    \end{itemize}
%The term ``semantically important'' allows the annotators to bring in their knowledge of the songs' cultural background.

%Each word in each song thus received one of three values: \textit{not\_important}, \textit{correct}, or \textit{wrong}, which the latter two also imply \textit{important}. 
%As a hint for distinguishing between \textit{important} and \textit{not\_important} words, we highlighted nouns, adjectives, verbs, adverbs and proper names based on POS tags obtained via the UDPipe tool \cite{straka_udpipe_2018}.
%To aggregate these values, we use \textbf{correctness}, defined as proportion of \textit{important} words marked as \textit{correct}.
%$$ correctness = \frac{\# important\_words\_labeled\_as\_correct}{\# important\_words} . $$

Annotators were shown songs and their translations via a dedicated web application,\footnote{Application source code, including detailed instructions: \url{https://anonymous.4open.science/r/folksong_mt_app-0685/README.md}} labelling each word in the source text as: (i) semantically important and correctly represented in the translation; (ii) semantically important but incorrectly rendered or absent; or (iii) not important for meaning. 
The term \textit{semantically important} is deliberately vague, allowing annotators to draw on their cultural backgrounds.
%knowledge of the songs. 
These word-level labels (\textit{correct}, \textit{wrong}, \textit{not\_important}) are aggregated into a single song-level metric, \textbf{correctness}: the proportion of \textit{important} words marked as \textit{correct}.
While MT literature offers a range of more complex metrics, correctness is sufficient and readily interpretable for our purpose of assessing translation adequacy as a preprocessing step for topic modeling.

%For each \textbf{song}, annotators were further asked: %presented with three overall questions:
%\begin{itemize}[noitemsep,nosep,leftmargin=*]
%    \item Overall rating: 0 to 5 points
%    \item Hallucinations: yes/no hallucinated (content-bearing) words present 
%    \item Unusual language: level of presence of unusual language in the song (none, barely, significant)
%\end{itemize}

At the song level, annotators additionally provided an overall quality rating (0--5); a binary judgement on the presence of hallucinated content-bearing words; and a three-point scale for the degree of non-standard language in the source (none, barely, significant).

%\noindent
%The web application source code, including detailed instructions, is part of the supplementary material.\footnote{\url{https://anonymous.4open.science/r/folksong_mt_app-0685/README.md}}

\subsection{Topic modeling}
\label{subsec:modeling}

Topic modeling has a history of over two decades in NLP, and its application to song lyrics is well-established: Latent Dirichlet Allocation (LDA) has been applied to popular music lyrics across typologically diverse languages including English \cite{hunke_evolution_2025}, Japanese \cite{sasaki_lyricsradar_2014}, Manipuri \cite{devi_exploiting_2020}, and Indonesian \cite{laoh_indonesians_2018}, as well as to folk song lyrics in Ukrainian \cite{petrovych_topic_2024} and Slovenian \cite{strle_language_2016}. 
However, LDA does not leverage the full range of word embeddings and thus relies heavily on dataset size.
We adopt a neural approach, which has demonstrated utility in related work on folkloristic material \cite{tangherlini_travels_2024} and, crucially, allows the use of a pretrained model, %computed on a large general-purpose reference corpus
%and thus 
compensating for the limited size of our dataset. %while providing a ready-made topic vocabulary, allowing direct comparison across songs regardless of their language of origin. 

%In our study, we are testing the argument 
%of loss of information in the projection to a shared space of translation does not necessarily imply loss of information relevant to the analytical method
%by applying a topic modeling pipeline to a multilingual corpus that was machine-translated into English. 
%We treat each song as a document and represent it as a distribution over a fixed set of topics.
%
%Because the folksong data itself is not large enough to train a reliable topic model, we use one computed on a large general-purpose reference corpus, 

We represent each song as a probability distribution over a fixed set of topics.
We use the pretrained BERTopic model \cite{grootendorst2022bertopic} MaartenGr/BERTopic\_Wikipedia,\footnote{\href{https://huggingface.co/MaartenGr/BERTopic\_Wikipedia}{\texttt{huggingface.co/MaartenGr/BERTopic\_Wikipedia}}} which was trained on the first paragraphs of approximately one million Wikipedia articles.
%, to each document via the \texttt{approximate\_distribution} method. 
This yields a vector of soft topic membership probabilities over the full topic vocabulary (2376 topics). 
The Wikipedia-based model was selected as the closest available approximation of a neutral, general-purpose topic model, minimising additional bias.
%introduced by the choice of topic model.
%beyond those already inherent in the machine translation step. 
%Applying this pipeline to our multilingual corpus, which was machine-translated into English prior to analysis, then allows direct cross-lingual comparison of lyrical content. 
%
%This approach deliberately sacrifices fine-grained linguistic detail in favour of a simple language-neutral representation, but we assume that the information loss introduced by translation does not necessarily entail loss of information relevant to topic-level analysis.

\subsection{Classification experiments}

%The classification performance thereby serves as an indirect

%To measure the extent to which the song lyrics are systematically associated with the social contexts in which those songs are used, we train classifiers with the processed lyrics of each song %(represented as a probability distribution over the 2367 topics of the BERTopic model) 
%as inputs and the typological class membership (labels) as the targets.
We train classifiers using the topic vectors derived from lyrics of each song as input and typological class membership (label) as the target, measuring how much the content of lyrics is systematically associated with social context. 

Higher scores are not necessarily better here: it may be that no appreciable relationship exists. 
%However, information can also be lost during the text representation (translation and topic modeling). Another source of artificially low scores may be issues in the classification setup.
%itself. 
The results should be taken as a lower bound, though, as the translation and topic modeling steps may discard information. %on how much might be found with a better representation. 
If the models are highly accurate for at least some labels, then the chance that classification results are artificially low because of bad input representations 
%have lost a lot of information 
does, in principle, decrease (though this is hard to measure): in this sense, the actual classification results we achieve are a baseline to be improved upon, though with an unknown upper bound which itself is the object of interest.
%employed machine learning and prepared an experiment in which songs are classified into typological categories derived from the labels provided by the respective corpora.

% Concretely, we train a classifier on the lyrical content of each song, represented as a probability distribution over BERTopic topics (vectors of length 2367) with typological class membership (labels) as a target. The classification performance thereby serves as an indirect measure of the extent to which the content of song lyrics is systematically associated with the social contexts in which those songs are used.

We perform classification experiments both on the single-language corpora, and the combined weddings corpus.
Evaluation of each model was performed using a stratified five-fold cross-validation scheme. 

We retain the 10 most frequent labels for each corpus.\footnote{Which labels are retained for each language and counts are given in the Supplementary materials: \url{https://anonymous.4open.science/r/ISMIR26_Folksong_weddings-B7F5}}
%
%Because even within the top 10 labels the class distribution is sometimes highly uneven, 
To further mitigate the influence of class imbalance, 
%the largest %overrepresented classes, 
we subsample within each training fold the largest classes to the size of the 5th largest class (or the smallest class if fewer than five classes). %just under the largest gap in the empirical class frequency distribution. 
%For wedding songs, we subsample to the smallest class size (Estonian). 
%By that we reacted to the uneven distribution of labels, where simply lowering the biggest class to the size of the second biggest can just create two big classes.

% TODO: How many dimensions retained per corpus? 
% - that is part of results is it not???
Also within each fold, 
%prior to running classification experiment, 
Principal Component Analysis (PCA) was run on the training data (topic probabilities per song), retaining components explaining 95\% of variance. %PCA was fitted exclusively on the training partition of each cross-validation fold and subsequently applied to the held-out validation partition.

% Should following paragraph be here or in results???
% this should go to results explaining the aggregated matrices
%To provide an aggregate view of model behaviour across folds, per-fold predictions were concatenated to form a single confusion matrix spanning the full training corpus, with each document contributing exactly once as a validation sample. 

We used Support Vector Machine (SVM) variants implemented in scikit-learn \cite{scikit-learn}: both a Linear SVM and a Radial Basis Function (RBF) kernel SVM, % and a Linear SVM. 
with Platt scaling \cite{guo_calibration_2017} and balanced class weighting.

% Code can be found in supplementary material LINK ...?

%Terminology:
%Topics = inferred content representation.
%Labels = the ``functional'' categories represented in folksong typologies.

%There are discrete and continuous representations. 
%Discrete = tokens (text in the original language or in the pivot language)
%Continuous = embeddings.

% ~~~~~~~~~~~~~~~~~~~~~~~~~~~~~~~~~~~~~~~~~~~~~~~~~~~~~~~~~~~~~~~~~~
\section{Results: Machine Translation}
\label{sec:mt}

% Short intro paragraph here?

\subsection{Coverage}

%As a shared representation for our multilingual corpora, we translated collected folksong lyrics into pivot language: English.

The translation failed for 5--15 \% of the song texts. The model returned \textit{``I cannot assist you with the request''} in the provided translation. Specifically, this affected 194/3320 CS songs, 419/3041 NL, 474/4515 ET, 1260/15861 KO, and 102/1764 UK songs. We were not able to diagnose and fix this issue (or issues), so we just discarded the songs that could not be translated from our datasets; this remains to be investigated in the future.

% \begin{itemize}[noitemsep,nosep]
%     \item CS: 194 out of 3320 songs
%     \item NL: 419 out of 3041
%     \item ET: 474 out of 4515
%     \item KO: 1260 ot of 15861
%     \item UK: 102 out of 1764
% \end{itemize}

% \subsection{Manual evaluation of MT: setup} %maybe better subsubsection ???
% \label{sec:translation:evaluation}

% MT results on folksong texts might not be as reliable as results on general translation tasks might suggest. While the downstream method is relatively coarse-grained and therefore robust to some MT error, the results of MT still must be inspected manually. 

\subsection{Manual evaluation of MT: results}

In Table \ref{tab:mt_eval} we present MT evaluation results. Czech, Dutch, and Ukrainian have a mean correctness of 0.9 and above, and a mean overall rating above 3; Korean and Estonian are noticeably worse. Also, Estonian and Korean reported 11/15 and 6/14 songs with hallucinations as well as orthographical anomalies in qualitative insights. 

% We introduce the metric \textit{correctness}, defined as:

% $$ correctness = \frac{\# important\_words\_labeled\_as\_correct}{\# important\_words} . $$
% While MT literature offers a range of evaluation metrics, our goal was to assess the viability of MT applied to folk song lyrics as a preprocessing step for the topic models, which require translation adequacy but not necessarily fluency. For this purpose, the proposed metric offers a sufficient and readily interpretable measure of translation quality.

% Korelace dialekt vs rating, individualita ratingu etc
Spearman $\rho$ 
%correlation coefficient 
between overall rating and correctness across all languages is a strong 0.7: correctness does not misrepresent overall translation quality too severely, though 
%as shown in \autoref{fig:eval_rating}, 
an overall rating of 3 shows a wide range of correctness (between 0.55 and 1.0).
%, and appear to be influenced by individual annotator background -- both linguistic and ethnomusicological. 
%
%The influence of unusual language varieties, primarily dialectical forms, can be observed in \autoref{fig:eval_language}.
Contrary to expectation, the presence of unusual language does not correlate much with lower correctness (Spearman $\rho$ of only $-$0.2).

\begin{table}[]
    \centering
    \begin{tabular}{rrccc}
    \toprule
         \textbf{Language} & \thead{\# words \\ evaluated} & \thead{Mean \\ importance} & \thead{Mean \\ correctness} &  \thead{Mean \\ overall} \\ % & \thead{Self\\agreement} \\
    \midrule
    %\midrule
         Czech & 764 & 0.84 & 0.97 & 3.67   \\
    %\hline
         Dutch & 1396 & 0.48 & 0.93 & 4.18  \\
    %\hline
         Estonian & 903 & 0.79 & 0.76 & 2.94  \\
    %\hline
         Korean & 584 & 0.95 & 0.79 & 3.00 \\
    %\hline
         Ukrainian & 1144 & 0.64 & 0.90 & 3.32  \\
    \bottomrule
    \end{tabular}
    \caption{MT manual evaluation results.}
    \label{tab:mt_eval}
\end{table}

\begin{comment}
\begin{figure}[t]
    \centering
    \includegraphics[width=1.0\linewidth]{latex/img/correctness_vs_rating_all_languages.png}
    \caption{Relation between the overall rating and correctness for each evaluated song. (Axis y does not start at 0.)}
    \label{fig:eval_rating}
\end{figure}
\end{comment}

\begin{comment}
\begin{figure}[t]
    \centering
    \includegraphics[width=1.0\linewidth]{latex/img/correctness_vs_unusualLanguage_all_languages.png}
    \caption{Overview of the relation between the annotated level of unusual language used and correctness for each evaluated song. Axes y does not start at 0.}
    \label{fig:eval_language}
\end{figure}
\end{comment}

% Self-agreement
%To assess annotator self-consistency, 
Each annotator was presented with one song twice, without their knowledge (one annotator spotted this and did not mark the song a second time). Of the six repeated annotations obtained, four yielded perfect agreement (Cohen's $\kappa$ 1.0). 
Once, a lower score was observed in the second annotation due to a larger set of words being marked as important, though the correctness of words flagged as important in both runs agreed. 
Once, broader disagreement across categories was seen, resulting in a Cohen's $\kappa$ of 0.69.
% How to deal with Inna saying "(see no. 3)"...? seventh annotation obtained
However, overall, correctness seems well-defined.

% IAA  - make it a section or at least make it bold
For Czech and Ukrainian each, an \textbf{extra complete annotation} was collected to measure %enable a preliminary  assessment of 
inter-annotator agreement (IAA). 
In Czech, the proportion of words labelled as important by both annotators with agreement on \textit{correct} or \textit{wrong} is 94.3\%. 
(A large difference in the proportion of words considered \textit{important} made using Cohen's kappa %effectively 
impossible.) % to effectively use % a standard IAA metric: Cohen's kappa. 
The mean absolute difference in overall translation quality between annotators is 0.7.
In Ukrainian, inter-annotator differences were more pronounced, with a mean correct/wrong agreement of 84\% on important words and a mean abs. difference in overall ratings of 1.3. This reflects the subjective nature of importance judgments, but agreement on correctness remains high.
%However, we 

%We also collected qualitative insights.
%
%The Czech annotators were surprised by the high quality of the translation for these specific texts and by the machine translation's general ability to handle the few dialects present. 

% Korean (DH)
% However, in some cases—due to strong dialectal variation—I was unable to determine the exact meaning even after consulting dialect dictionaries.
%
% Additionally, I noted that the terms “성seong” and “형hyeong” were marked as incorrect. Although these terms are typically used between males in modern Korean, in the context of these folk songs they refer to older female relatives, such as sisters or female cousins.
%
% When a line of lyrics is spoken by a character but omits the subject and object, it is sometimes rendered as a declarative sentence.\nFor example, “washing his face and eating the morning meal” can represent the speaker addressing her father-in-law, meaning: “Wash your face and eat your breakfast.”
% Words are written phonetically rather than following standard orthography, which makes translation more difficult. 
%
From qualitative insights into MT issues, we learned that for Estonian, some words are mistranslated although they are present in standard Estonian (e.g. words \emph{nadu} (husband's sister) and \emph{küdi} (husband's brother) -- probably because there is no precise English equivalent), meaning of some of the songs got lost, and the overall less satisfactory results may also be due to non-standard orthography used in the data, which was reported as challenging also in Korean.
A Ukrainian annotator stated that while the translation generally conveys the plot and basic semantics, it systematically loses features of the original, such as diminutive forms and traditional formulaic expressions, while culturally specific terms are often replaced by Western European analogues, and so often fails to reproduce ethnographic and symbolic specificity.
Dutch and Czech annotators had no negative comments.

% Comments from IL annotations:
% which is often inadequately rendered in automatic translation. While translation generally conveys the plot and basic semantics, it systematically loses key features of the original, including formulaic repetition, affective meaning encoded in diminutive forms  \emph{kalynonka}(little viburnum), and culturally specific terms, which are often replaced by Western European analogues \emph{pan}(Lord). In particular, ritual vocabulary is frequently reduced \emph{korovai}(wedding bread), resulting in the loss of ethnographic specificity and symbolic meaning.

% The model handles basic semantics well but loses formulaic structure and fails to reproduce ritual language.
% stylistic inaccuracies
% The overall meaning is conveyed, but the folkloric style is partially lost.
% Diminutive forms have been replaced with neutral ones
% The translation partially conveys the meaning, but contains several inaccuracies and loses important folkloric features: formulaic repetition, rhythm, and dialectal elements are weakened, and some lexical choices (“growl”).
% Loss of folkloric formulaicity
% The translation conveys the overall narrative and structure, but partially loses the folkloric style and ritual functions

\section{Results: Lyrics vs. Typology}
\label{sec:results_class}
%Topic models in pivot language}

\begin{comment}
\begin{table*}[]
    \centering
    \begin{tabular}{|l|r|r|r|r|r|r|r|}
    \hline
          & cs & nl & et & ko & uk & weddings & \thead{cs, uk, nl \\ weddings}\\
    \hline     
         avg. weighted F1-score & 0.258 & 0.659 & 0.525 & 0.638  & 0.456 & 0.699 & 0.645 \\
    \hline
         avg. \# PCA components & 419 & 377.2 & 472.2 & 494 & 236.4 & 284 & 186.2 \\
    \hline

    \end{tabular}
    \caption{RBF classif results and PCA reduction avg on individual corpora sets and on two wedding subsets}
    \label{tab:class_eval}
\end{table*}
\end{comment}

\begin{comment}
% Table with results for "largest gap" subsampling
\begin{table}[t]
    \centering
    \begin{tabular}{r|ccc}
         \textbf{Corpus} & \thead{Avg. weighted \\ F1 score} & \thead{Avg. # PCA \\ components} & no. of songs \\
         \hline
         Czech & 0.26 & 419 & 2829\\
         Dutch & 0.66 & 377 & 2612 \\
         Estonian & 0.53 & 472 & 3553 \\
         Korean & 0.64 & 494 & 7767 \\
         Ukrainian & 0.46 & 236 & 1585 \\
         \hline
         Weddings & 0.70 & 284 & 1197 \\
         \thead{Weddings \\ CS,NL,UK} & 0.65 & 186 & 617
    \end{tabular}
    \caption{Classification results on monolingual corpora and on wedding songs (all languages, and the 3 well-translated ones): avg. weighted F1 score, and avg. number of PCA components explaining 95~\% of embeddings variance.}
    \label{tab:class_eval}
\end{table}
\end{comment}

\begin{table}[t]
    \centering
    \begin{tabular}{rccc}
        \toprule
         \textbf{Corpus} & \thead{Avg. weighted \\ F1 score} & \thead{Avg. \# PCA \\ components} & \thead{no. effective \\ training songs} \\
        \midrule
        \midrule
         Czech & 0.26 & 411 & 1685 \\
         Dutch & 0.53 & 230 & 499 \\
         Estonian & 0.51 & 461 & 2415 \\
         Korean & 0.63 & 488 & 3552 \\
         Ukrainian & 0.52 & 293 & 1040 \\
         \midrule
         Weddings & 0.63 & 216 & 500 \\
         \thead{Weddings \\ CS,NL,UK} & 0.65 & 186 & 447 \\
        \bottomrule
    \end{tabular}
    \caption{Classification results on monolingual corpora and on wedding songs (all languages, and the 3 well-translated ones): avg. weighted F1 score, and avg. number of PCA components explaining 95~\% of the embeddings variance, and the number of training songs after subsampling.}
    \label{tab:class_eval}
\end{table}

\begin{figure*}
    \centering
    \includegraphics[width=1.0\linewidth]{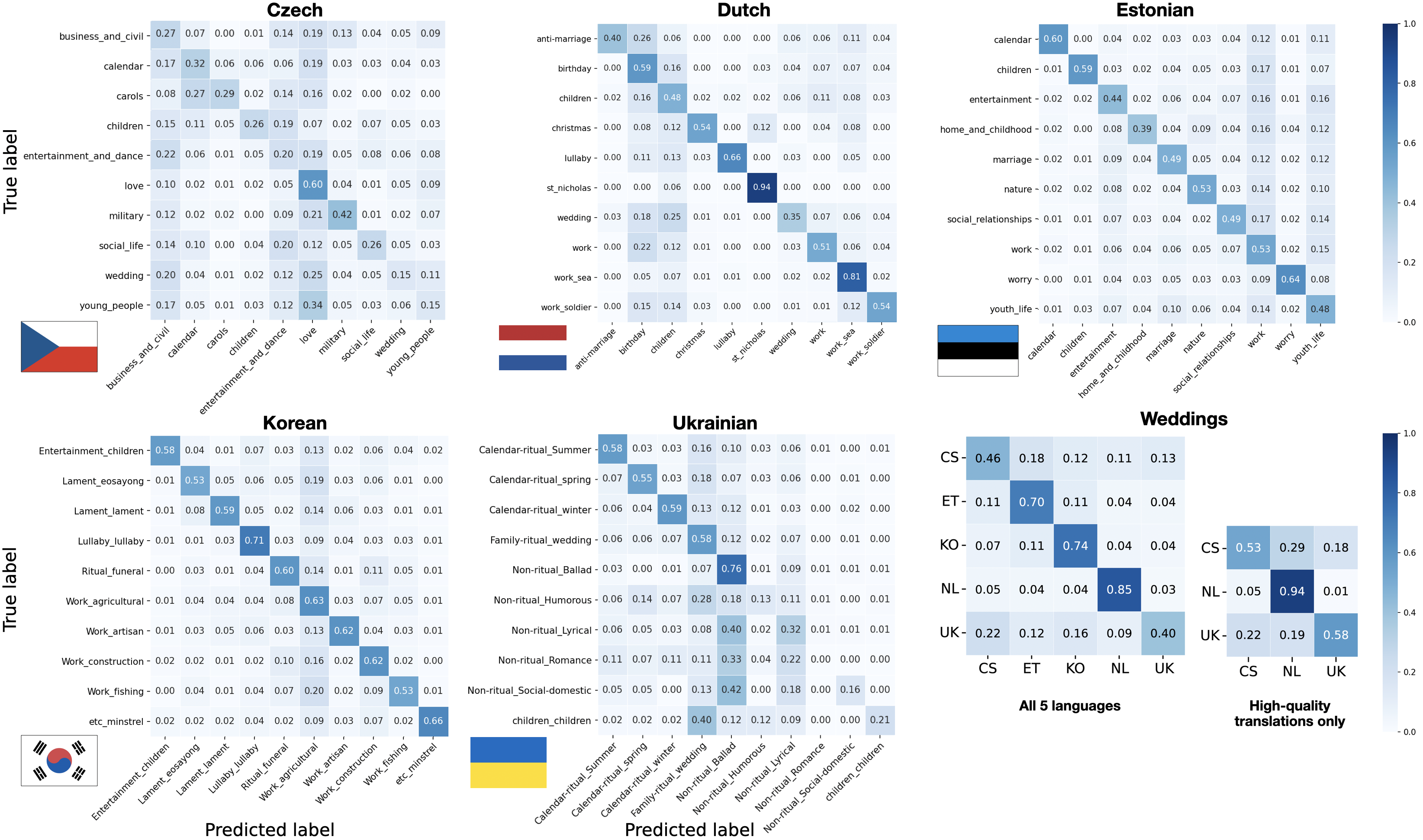}
    \caption{Classification results: row-normalised confusion matrices for top 10 labels in each language, and on the languages for wedding songs. We report recall because we care most about the question: ``Is this label hard to identify?''.}
    \label{fig:matrices}
\end{figure*}

%\textbf{Preprocessing.} - DESRIBED IN METHOD
%We discard songs that have typological labels from multiple categories that we study.

%Only use most frequent 10 topics for each language.

%Also discard all "Other", "unknown", and other labels that express uncertainty and records with missing labels.

%Subsample large topics.

%PCA on train part of 

%As described in \autoref{subsec:modeling} 

% \textbf{BERTopic} % TODO: Move higher?
% Neutral language, ``matter-of-fact'' descriptive language, so good for ``what things are there'' --- what objects and concepts are people singing about.

\subsection{Classification: monolingual corpora}

% For all five languages we restrict our data sets to the ten most frequent labels. On such subsets, we train two classifiers (SVM with RBF kernel and linear SVM) targeting these labels (typologies of respective corpora). 

% In all cases, the RBF kernel performs better than linear SVM in terms of F1 score; hence we only report results of the RBF SVM. \autoref{tab:class_eval} shows the mean of weighted F1 scores for different data sets as well as the average number of PCA components.

Tab.~\ref{tab:class_eval} shows the mean of weighted F1 scores for different data sets as well as the average number of PCA components.
In all cases, the RBF kernel performs better than linear SVM in terms of F1 score; hence we only report results of the RBF SVM. Fig.~\ref{fig:matrices} shows %row-normalised 
confusion matrices.
Correlation between effective training dataset size (after subsampling) and aggregated F-score is low (0.32).\footnote{Spearman $\rho$ between label size and F-score across individual corpora averages <0.25 regardless of whether we consider effective training size or just label size.}

\subsection{Cross-Cultural Classification: Wedding songs}
\label{sec:weddings}

%From the gathered data we picked songs with wedding related label in the typology as a well defined set (compared to other picking other labels that might not be represented in all corpora or might have more problematic alignment). As mentioned in \ref{subsec:ko} Korean situation in this is a bit different and what is in the corpus regarding weddings is more of a lament about married life than songs regarding weddings. For that reason we look at this Korean subsample as at control group. 
%We consider all Estonian multi-topic songs that are also wedding songs to be wedding songs (16 songs).

%In this situation, classifiers were trying to predict the language of origin of song lyrics, as they all share the label (context) of a wedding. 
In this scenario, classifiers attempt to predict the language of origin of song lyrics (culture). The lower the score, the less distinct are the wedding song lyrics of the given corpus from wedding songs of other corpora.
The results are shown in Fig.~\ref{fig:matrices}, bottom right panel; the overall F-score is $0.65$ (see Tab.~\ref{tab:class_eval}).
We also re-ran classification with only the high-correctness languages (CS, NL and UK).
\section{Discussion and Conclusions}
\label{sec:conclusions}

%Conclusions we can make and why:

Machine translation is becoming good enough to open up this research space, if one is willing to give up nuance within individual cultures for the ability to make comparisons. Known issues with MT quality imbalance across languages \cite{romanou_include_multilingual_2024, jumelet_multilingual_blimp_2025} apply; however, reported folksong-specific issues with orthography may be the main driver of worse MT performance on ET and KO.

The strength of the relationship between the content of lyrics and social context varies greatly by language. For NL, St.~Nicholas and sea work songs are identified much better than anything else. Conversely, for CS, which also has high-quality translations, the relationship between lyrics content and social context is practically non-existent.\footnote{This may have been due to naive merging of labels, but without merging labels, the weighted F-score is 0.19, so the consolidation did not artificially damage classification. A more aggressive merging scheme also did not change the results noticeably.}
In UK, some relationship between content and context is found for the Calendar-ritual classes and ballads, while non-ritual songs show no relationship at all. For KO and ET, approx. half of the songs from each class are classified into their label (though given their worse translation results, this should be examined in more detail). There clearly is something to analyse here!

%3) The chosen multilingual representation pipeline (for the well-translated languages, at least) seems to allow for meaningful analysis of the relationship between lyrics and social contexts. 

Classifying wedding songs by culture seems easier than classification of songs by topic within cultures. 
The finding that there is more variation within than between musical cultures \cite{mehr_universality_2019} may hold for musical features, but not for lyrics. 
%choices within this social context, this might not hold. 
The Korean control group is not more distinct than Dutch songs: ``complaint'' songs may be grouped together with wedding songs in different typologies. 
Czech and Ukrainian are much less recognisable by lyrics --- but they are \textit{not} confused for each other, so points about Slavic-ness would be misleading (in line with musical diversity being independent on language and genetics \cite{passmore_global_2024}). The lower accuracy may instead have to do more with how broadly the wedding label were defined when the corpora were being collected: comparative investigation of the actual topics present in the wedding songs is needed in future work.
%Weddings according to subsampled to 100: CS and UK clearly less distinctive than ET, KO, NL, although the confusion is \textit{not} mutual.

%On the three-language version (with high-quality translation), suddenly both CS and UK get confused for NL more! This is not a matter of class imbalance: there are \textit{fewest} Dutch songs here (and the same pattern emerges also when we ran this without subsampling). Lower accuracies for Czech and Ukrainian would make sense if songs in these two Slavic were confused for each other, but their accuracy is just as much hurt by their songs being classified as Dutch. Why does NL act as a ``collector'' class here? What is going on?

This brings us to our \textbf{limitations}. First, the typologies for individual languages are far from compatible. Weddings were comparable by virtue of being such a strong anthropological universal, but further work must focus on better understanding how the social context labels have been assigned and how they relate to each other. The confusion between romances and ballads in UK may be because the labels were assigned rather subjectively. The genres also overlap thematically, particularly in their focus on love, separation, loss, and emotional suffering. While ballads tend to be more narrative and romances more emotion-oriented, this distinction may not always be evident from lexical content alone.

%This kind of depends on the results, doesn't it...
%What stood out is how difficult typologies of folksong are to compare. It is clear that the different ethnographic policies will have to be critically re-examined for future cross-cultural work.
%Assuming the methodology isn't fundamentally flawed in any of the steps, we could draw the following conclusions:

%\textbf{Single-language:} Doesn't classify very well. The same concepts (expressed via choice of words in lyrics) are re-used across different social contexts.
%(Possibly interesting: concepts that are NOT used...)
% Also, for some labels in some languages, there \textit{is} a very high score. So it's not that the topic model is uniformly bad.

%There can be overlap between categories of different granularity, e.g. 'carols' can be in different collection also part of 'calendar'. While our policy to labels was rather 'downstream', this would require manual inspection.

%Ukrainian: non-ritual labels get confused a lot more than ritual. Stability?

%Dutch: some labels identified really, really well. Others -- not so much.
% The confusion between christmas and birthday is gone...

%Estonian, Korean: No label has a very high score. This might be due to translation issues, where, despite a positive Spearman correlation for Estonian, 

% \textbf{Weddings:} there is some notion of similarity.
% Estonians get confused with Korean -- might be due to translation issues, as MT performed noticeably worse for both.
% However, Czech wedding songs get confused for Korean almost as much.

While the typological labels of songs were typically mutually exclusive, this was not the case for Estonian and Dutch, and this leads to a broader question: how legitimate is the categorisation of songs by social function? Songs may certainly serve multiple functions. The question of categorising vs. ``tagging'' songs should be revisited.

%Also, merging. %czech labels ...?

%Overall quality of the texts (cleaning the lyrics field to be only lyrics... no notes etc) 
%valiabilty vs usability of data?

A limitation may also be the BERTopic trained on Wikipedia being \textit{too} neutral/also specific style language; fine-tuning the topic model should be explored. % with folk lyrics data

The clear next step is to analyse in more detail the distribution wedding of songs in the lyrics representation space, and relate them to individual topics. This is (unfortunately) beyond the scope of this paper; however, all song representations including topic assignments are made available in Supplementary materials.

%\textbf{Conclusions.}
% JH: I really enjoyed formulating this last paragraph. :)
% Probably should be stated in more particular maner - we did THIS and THIS ...
The cross-cultural comparison of folksong lyrics of course cannot be resolved in a single paper. 
However, we hope that this work will serve as an opening shot. 
Perhaps more valuable than any specific result is exposing the issues that further research in this direction must address: specificities of folksong language, issues in understanding and harmonising ethnomusicological typologies, and more work on the methodological fundamentals of what kinds of insights can cross-cultural comparison of lyrics meaningfully bring. 
But given the importance of music and \textit{what we sing} to how we construct our collective identities, we believe these are questions worth looking into, and the technological tools may now be there for large-scale corpus-based inquiry that is at home in MIR to seriously contribute to this fascinating layer of human culture.

% FUTURE WORK gdě ???
% Aha... nad conclusions ... 

\section{Acknowledgments}

This work has been funded by the project “Human-centred AI for a Sustainable and Adaptive Society” (reg. no. CZ.02.01.01/00/23\_025/0008691), co-funded by the European Union.

It has also been partially supported by the project TK215 ``Estonian Roots: Centre of Excellence for transdisciplinary studies on ethnogenesis and cultural diversity''.

We extend our thanks to Martin Popel and Ondřej Bojar for helpful discussions on MT evaluation design.

\section{AI Usage Statement}

%In line with the ISMIR 2026 AI Usage Policy, if you need to declare and explain the use of AI tools in the work described in the paper, you can do so by including a section titled ``AI Usage Statement" at the end of the paper. The AI Usage Statement section can be included both at submission time and in your camera-ready version. See the Call for Papers and Author Guidelines for more details. The AI Usage Statement section does \textit{not} count towards the page limit for scientific content. 
GitHub Copilot was used to assist in coding the online interface for evaluating manual annotations. 
Grammarly was used to improve the text.

\bibliographystyle{unsrtnat}
\bibliography{bibliography.bib}

\newpage

\end{document}